\documentclass[10pt,twocolumn,letterpaper]{article}

\usepackage[final]{iccv}
\usepackage{graphicx}
\usepackage{multirow}
\usepackage{xcolor}

\definecolor{iccvblue}{rgb}{0.21,0.49,0.74}
\usepackage[pagebackref,breaklinks,colorlinks,allcolors=iccvblue]{hyperref}

\title{TeethGenerator: A two-stage framework for paired pre- and post-orthodontic 3D dental data generation
}

\author{
Changsong Lei\textsuperscript{1} \quad Yaqian Liang\textsuperscript{1} \quad Shaofeng Wang\textsuperscript{2} \quad Jiajia Dai\textsuperscript{1} \quad Yong-Jin Liu\textsuperscript{1} \\
\textsuperscript{1}Department of Computer Science and Technology, Tsinghua University, Beijing, China \\
\textsuperscript{2}Beijing Stomatological Hospital, Capital Medical University, Beijing,  China \\
{\tt\small leics23@mails.tsinghua.edu.cn} \quad{\tt\small yaqianliang@tsinghua.edu.cn}\quad {\tt\small 2939108747@ccmu.edu.cn} \\ {\tt\small \{daijiajia, liuyongjin\}@tsinghua.edu.cn} 
}

\begin{document}
\maketitle

\begin{abstract}
Digital orthodontics represents a prominent and critical application of computer vision technology 
in the medical field. So far, the labor-intensive process of collecting clinical data,
particularly in acquiring paired 3D orthodontic teeth models, constitutes a crucial bottleneck for developing tooth arrangement neural networks.   
Although numerous general 3D shape generation methods have been proposed,
most of them focus on single-object generation and are insufficient for generating anatomically structured teeth models,  each comprising 24-32 segmented teeth.
In this paper, we propose TeethGenerator, a novel two-stage framework designed to synthesize paired 3D teeth models pre- and post-orthodontic,  aiming to facilitate the training of downstream tooth arrangement networks. 
Specifically, our approach consists of two key modules: (1) a teeth shape generation module that leverages a diffusion model to learn the distribution of morphological characteristics of teeth, enabling the generation of diverse post-orthodontic teeth models; and (2) a teeth style generation module that synthesizes corresponding pre-orthodontic teeth models by incorporating desired styles as conditional inputs.
Extensive qualitative and quantitative experiments demonstrate that our synthetic dataset aligns closely with the distribution of real orthodontic data, and promotes tooth alignment performance significantly  when combined with real data for training.
The code and dataset are available at \href{https://github.com/lcshhh/teeth_generator}{https://github.com/lcshhh/teeth\_generator}.

\end{abstract}

\section{Introduction}
Computer vision technology has played an important role in many natural science and engineering fields, including digital orthodontics within the medical domain. Orthodontic treatment aims to correct irregular tooth arrangements to achieve an aesthetically pleasing, balanced, and stable dental arch with proper occlusal relationship. In digital orthodontics, designing ideal target tooth positions is a pivotal challenge, as these positions fundamentally influence treatment planning and clinical outcomes. Traditionally, this process relies heavily on the expertise and aesthetic judgment of dentists and technicians.
To enhance the efficiency and intelligence of digital orthodontic workflows, researchers have developed automated systems for predicting orthodontic target positions, e.g.,  TANet~\cite{wei2020tanet}, OrthoGAN~\cite{shen2022orthogan}, Tooth Motion Diffusion~\cite{fan2024collaborative}, TADPM~\cite{lei2024automatic}, etc. These data-driven approaches leverage insights from historical orthodontic cases to generate clinically reasonable target tooth positions for new patients, providing valuable decision-making support to orthodontists.



\begin{figure}
    \centering
    \includegraphics[width=0.8\linewidth]{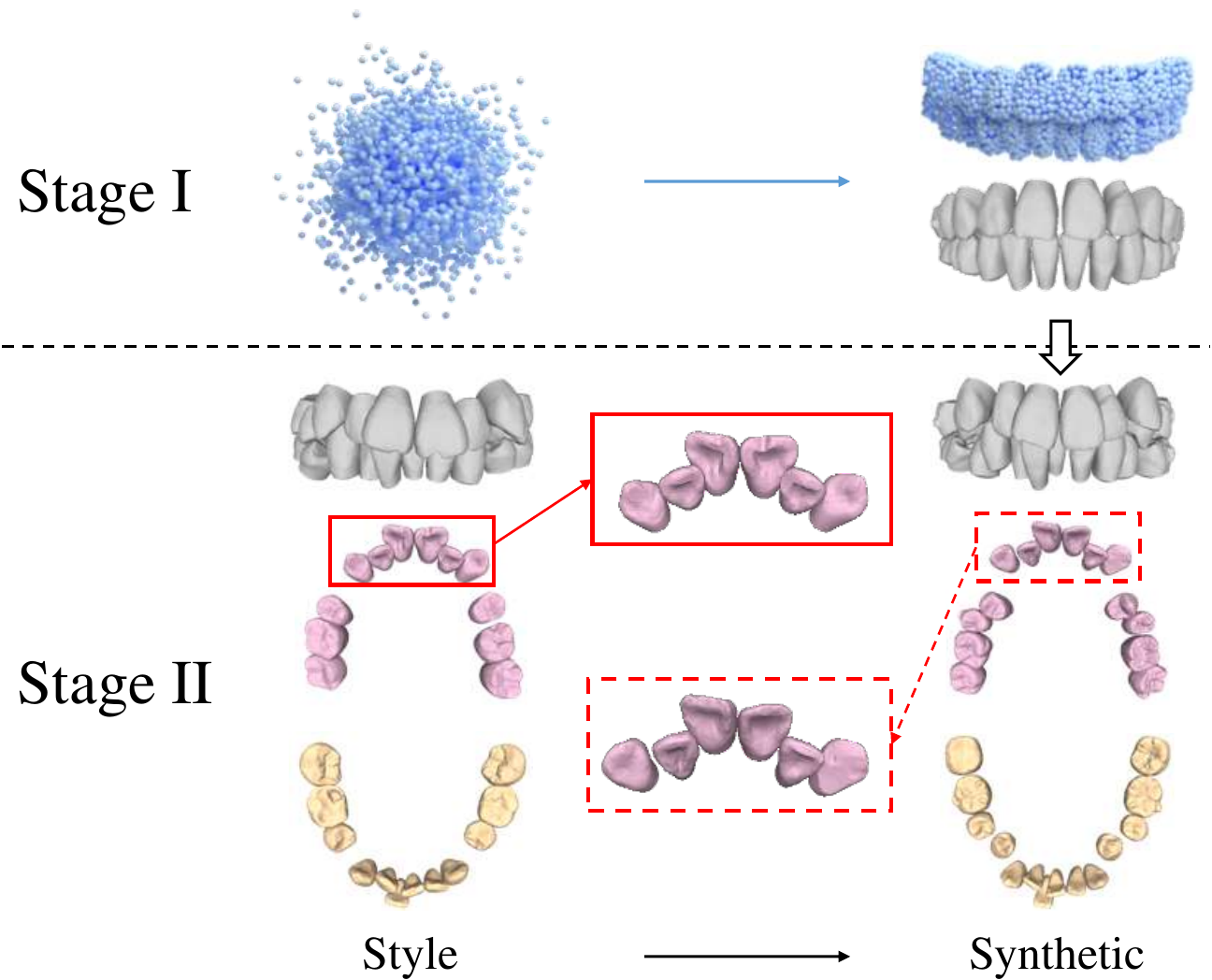}
    \caption{Illustration of the proposed TeethGenerator, where Stage I generates post-orthodontic teeth models and Stage II generates pre-orthodontic teeth models that have similar postures to the style model while preserving teeth's morphological identity.}
    \label{fig:first}
\end{figure}



All existing tooth alignment methods require substantial quantities of paired pre- and post-orthodontic 3D teeth models to train neural networks. 
However, acquiring such paired datasets remains labor-intensive and time-consuming, primarily due to that pre- and post-orthodontic teeth models must be scanned from real patients undergoing multi-year treatments. 
Furthermore, collecting teeth datasets raises concerns related to patient privacy and commercial issues, further complicating data acquisition.
Despite remarkable progress in deep learning-based tooth alignment, 
existing approaches~\cite{wei2020tanet,shen2022orthogan,fan2024collaborative,lei2024automatic} primarily validate their effectiveness on proprietary datasets, raising concerns about methodological generalizability. 
To date, only one publicly accessible 3D orthodontic dataset~\cite{Teethdataset} has been proposed, which still suffers from an uneven distribution of malocclusion categories. 
For instance, it contains only 20 samples of anterior open bite.
This critical data scarcity severely constrains progress in intelligent digital orthodontics.
To overcome these data limitations, this paper aims to generate high-quality synthetic paired 3D teeth models and explore whether the synthetic data can improve the performance of tooth arrangement neural networks.


3D shape synthesis has long been a hot research topic in computer vision and graphics, with foundational architectures including GAN-based frameworks~\cite{chen2021decor,wen2021learning}, VAE-based approaches~\cite{tan2018variational}, and recent diffusion-based 3D paradigms~\cite{yang20233dstyle,raj2023dreambooth3d,jain2022zero,pooledreamfusion,lin2023magic3d}. 
Despite these advancements in general 3D generation, their applicability to 3D teeth model synthesis remains limited due to domain-specific complexities.
%
In particular, generating 3D teeth models involves strong medical priors, including:
(1) \textbf{Multi-instance generation}: unlike conventional single-object synthesis, 3D teeth models require the simultaneous generation of 24–32 segmented and tightly integrated teeth point clouds.
(2) \textbf{Distribution matching}: the postures and alignments of teeth in generated teeth models must match the distribution of real data. 
(3) \textbf{Orthodontic tooth consistency}: 
to facilitate the training of subsequent teeth arrangement networks, the shape and size of each corresponding tooth in synthesized pre- and post-orthodontic models must be strictly consistent.
(4) \textbf{Versatility of teeth posture}:
the generated 3D teeth models should  exhibit stylistic diversity, including rare cases, and avoid repetition of common case types. 
These requirements collectively lead to the teeth data synthesis becoming a distinct and challenging problem.

To address the above challenges, we propose TeethGenerator, a novel framework that synthesizes 3D teeth models before and after orthodontic treatment through a two-stage process.
In Stage I, we synthesize post-treatment teeth models using diffusion models. To achieve this, we design a specialized feature extractor based on a Vector Quantized-Variational Autoencoder (VQ-VAE)~\cite{VQ_VAE} to compress 3D tooth point clouds into discrete latent representations. 
The diffusion model then learns the distribution of this latent space. By decoding the latent vectors sampled by the diffusion model, our method can generate post-orthodontic teeth models with various tooth morphologies.
In Stage II, we employ a Transformer architecture to synthesize pre-orthodontic teeth models according to the style model. Specifically, we extract style information from the given style model and shape information from the post-orthodontic teeth models using the proposed feature extractors. The extracted style information is then fed into the Transformer as input, while the shape information is incorporated into the attention mechanism as a conditioning factor to generate the corresponding tooth transformation parameters. By applying these parameters to the post-orthodontic teeth models generated in Stage I, we obtain the synthesized pre-orthodontic teeth model, while ensuring the size and shape of each tooth remains identical.

The main contributions of this paper are as follows:
\begin{itemize}
 \item We propose the first pair-wise teeth generation framework  capable of generating paired pre- and post-orthodontic 3D teeth models with diverse styles, consistent with the true distribution of real teeth data.



\item 
We develop a two-stage data synthesis framework.
In Stage I, we combine VQ-VAE and diffusion models to learn the distribution of real tooth shapes and alignments, enabling the generation of post-orthodontic tooth point clouds. In Stage II, we generate transformation parameters based on the desired style to obtain pre-orthodontic tooth point clouds while preserving tooth morphologies.



\item Extensive experiments demonstrate the effectiveness of our synthetic teeth data. The generated data not only improves tooth alignment performance when used to train neural networks, but also has the potential to assist in the training of orthodontists and technicians.
\end{itemize}
 
\section{Related Works}

\subsection{Synthetic Data for AI Training}

With the  development of AI algorithms and network structures, significant progress has been made in artificial intelligence-generated content. In the field of dentistry, ~\cite{hwang2018learning} used GANs to generate dental crowns that exceed human technicians in quality. Recently, denoising diffusion probabilistic models (DDPMs) have emerged as a new category of generative techniques that are capable of generating high-quality synthetic data across a variety of domains, e.g., images~\cite{saharia2022palette}, text~\cite{brown2020language}, and videos~\cite{Ho2022video}.  


\begin{figure*}[t]
    \centering
    \includegraphics[width=1\linewidth]{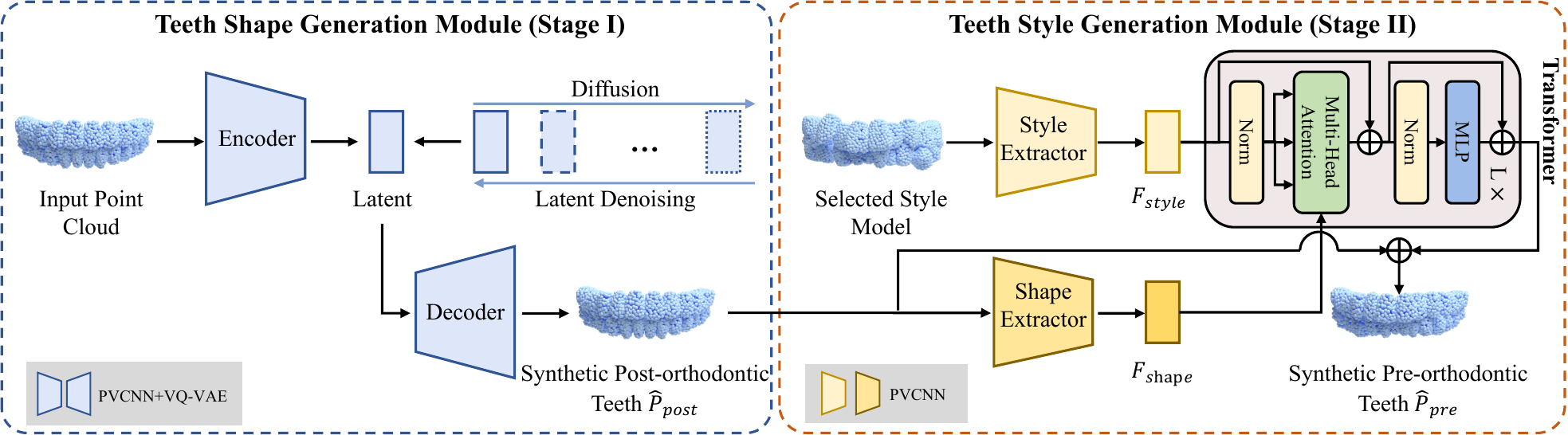}
    \caption{
    The pipeline of the proposed TeethGenerator. 
    In Stage I, we first train a VQ-VAE to reconstruct the teeth point clouds. Then, we utilize a diffusion model to sample latent encodings in the  latent space, which are subsequently decoded by the VQ-VAE decoder into post-orthodontic teeth models. 
    In Stage II, we extract the style information from the selected style model and extract the shape information from the post-orthodontic teeth models, respectively. Then, we employ a Transformer framework and incorporate style information and shape information to guide the transformation parameters prediction, enabling the generation of pre-orthodontic teeth models with the desired style. Note that all the feature extractors in our framework are built on PVCNN.
    }
    \label{fig:architecture}
\end{figure*}

As there are more and more synthetic data generated, it is increasingly valuable to explore its potential for training AI networks. Studies such as DCFace~\cite{kim2023dcface} and \cite{azizi2023synthetic} have demonstrated that using high-quality synthetic data to train networks can enhance the performance of vision recognition tasks. While some research focuses on whether synthetic data can effectively train generative networks, Shumailov et al.~\cite{shumailov2024ai} investigated the recursive training of large language models using the generated text. They found that AI models tend to collapse when trained on recursively generated data.
Our goal of this paper is to generate paired pre- and post-orthodontic teeth models and explore the effectiveness of using synthetic 3D data to train neural networks of downstream tooth arrangement tasks.



\subsection{3D Shape Generation}

Previous 3D  generation methods build on various frameworks, e.g., generative adversarial networks (GANs)~\cite{chen2021decor,wen2021learning}
variational autoencoders (VAEs)~\cite{tan2018variational}, autoregressive models~\cite{nash2020polygen}, and so on. 
Currently, along with the emergence of denoising diffusion models (DDMs),
3D shape generation has become an attractive research area.
A growing number of papers, such as DreamField~\cite{jain2022zero}, DreamFusion~\cite{pooledreamfusion}, and Magic3D~\cite{lin2023magic3d} leverage prior knowledge of 2D image diffusion models or vision language models to create 3D models from text or images. Many researchers generate multi-view images first to promote efficiency and geometric consistency of shape generation.  Such as One-2-3-45~\cite{liu2024one} generates multi-view images for the input view and then lifts them to 3D space, while Wonder3D~\cite{long2024wonder3d} generates multi-view consistent normal maps and their corresponding color images with a cross-domain diffusion model. There exists another group of works that attempt to train 3D generative diffusion models to directly produce 3D point clouds~\cite{luo2021diffusion,zhou20213d} or 3D meshes~\cite{liumeshdiffusion,siddiqui2024meshgpt}. For example,  LION~\cite{vahdat2022lion}introduces the hierarchical latent diffusion models for 3D shape generation from the hierarchical latent space. 
Most of the above methods generate single 3D models once. In contrast, this paper aims to generate paired upper and lower teeth models.

\subsection{3D Teeth Models Arrangement}

At present, there is no related research on generating synthetic pre-orthodontic 3D teeth models. 
In this paper, we propose to simulate pre-orthodontic teeth by predicting transformation parameters that adjust the post-treatment teeth models back to non-alignment. 
This approach is conceptually similar to predicting orthodontic target positions but in reverse. 
Early approaches to modeling teeth arrangement involve extracting feature representations from original 3D teeth models and then using a multi-layer perceptron (MLP) to predict teeth transformation parameters directly. 
For instance, TANet~\citep{wei2020tanet}  employs PointNet to extract the features of the teeth and implement feature propagation through topological relations with graph neural network, PSTN~\citep{li2020malocclusion} arranges teeth by leveraging the spatial interrelationship between different teeth, and \citet{wang2022tooth} propose to arrangement tooth automatically in a hierarchical graph structure. Rather, other studies focus on generating images of teeth after orthodontic treatment.  TAligNet~\citep{lingchen2020iorthopredictor} generates the facial image with aligned teeth, simulating a real orthodontic treatment effect.  OrthoGAN~\citep{shen2022orthogan}  progressively generates the patients' frontal facial images based on transfer learning GAN.
More recently, TADPM~\cite{lei2024automatic}  proposed to introduce the DDPMs to learn the distribution of teeth transformation matrices.  Similarly, Fan et. al.~\cite{fan2024collaborative} leverage the diffusion model to learn the distribution of tooth motions throughout the orthodontic process.


\section{Method}
We propose TeethGenerator, a two-stage data synthetic framework for generating 3D teeth point clouds. The overall architecture of our method is shown in Fig.~\ref{fig:architecture}, including two distinct modules to generate pre- and post-orthodontic 3D teeth models, respectively.  In Stage I, we propose the Teeth Shape Generation Module to generate \textbf{post-orthodontic} 3D teeth models with diverse tooth morphologies. 
In Stage II, we propose the Teeth Style Generation Module to generate corresponding \textbf{pre-orthodontic} 3D teeth models according to the given teeth arrangements style.
Note that the actual output of our method is point clouds, while meshes are converted from point clouds only for visualization.

\subsection{Preliminary}

Diffusion probabilistic models (DPMs) are a class of generative models that gradually convert a Gaussian distribution into the target data distribution through an iterative denoising process.
%
Let $\mathbf x_0\in\mathbb R^D$ be a $D$-dimensional random variable with an unknown distribution $q_0(\mathbf x_0)$.
DPMs define a forward diffusion process where the information of $\mathbf x_0$ is progressively corrupted by gaussian noises, such that for any 
$t\in[0,T]$, the transition distribution ~\citep{song2020score,sohl2015deep} can be formulated as:
\begin{align}\label{eq:forward}
q_{0t}(\mathbf x_t|\mathbf x_0)=\mathcal N(\mathbf x_t;\alpha_t\mathbf x_0,\sigma_t^2\mathbf I),
\end{align}
where $\alpha_t$ and $\sigma_t>0$ are functions of $t$ with bounded derivatives. 
The denoising process is achieved by training a denoiser network $\epsilon_\theta(\mathbf {x_t},t)$ to predict the initial noise $\epsilon$ using the following objective:
\begin{equation}\label{eq:diff_loss}
    \mathcal{L} = \mathbb{E}_{t, \mathbf{x}_0, \epsilon} \left\| \epsilon_\theta \left( \mathbf {x_t}, t \right) - \epsilon \right\|_2^2 .
\end{equation}

\subsection{Stage I: Teeth Shape Generation Module}
\label{voxel}


In this stage, we first train a VQ-VAE to encode the teeth point clouds into latent space and decode the latent vectors to reconstruct the aggregated teeth point clouds. Then we train the voxel-based diffusion model in the latent space to learn the distribution of shape and size characteristics of teeth models, enabling the generation of point clouds with varying morphologies.


\begin{figure}
    \centering
    \includegraphics[width=1\linewidth]{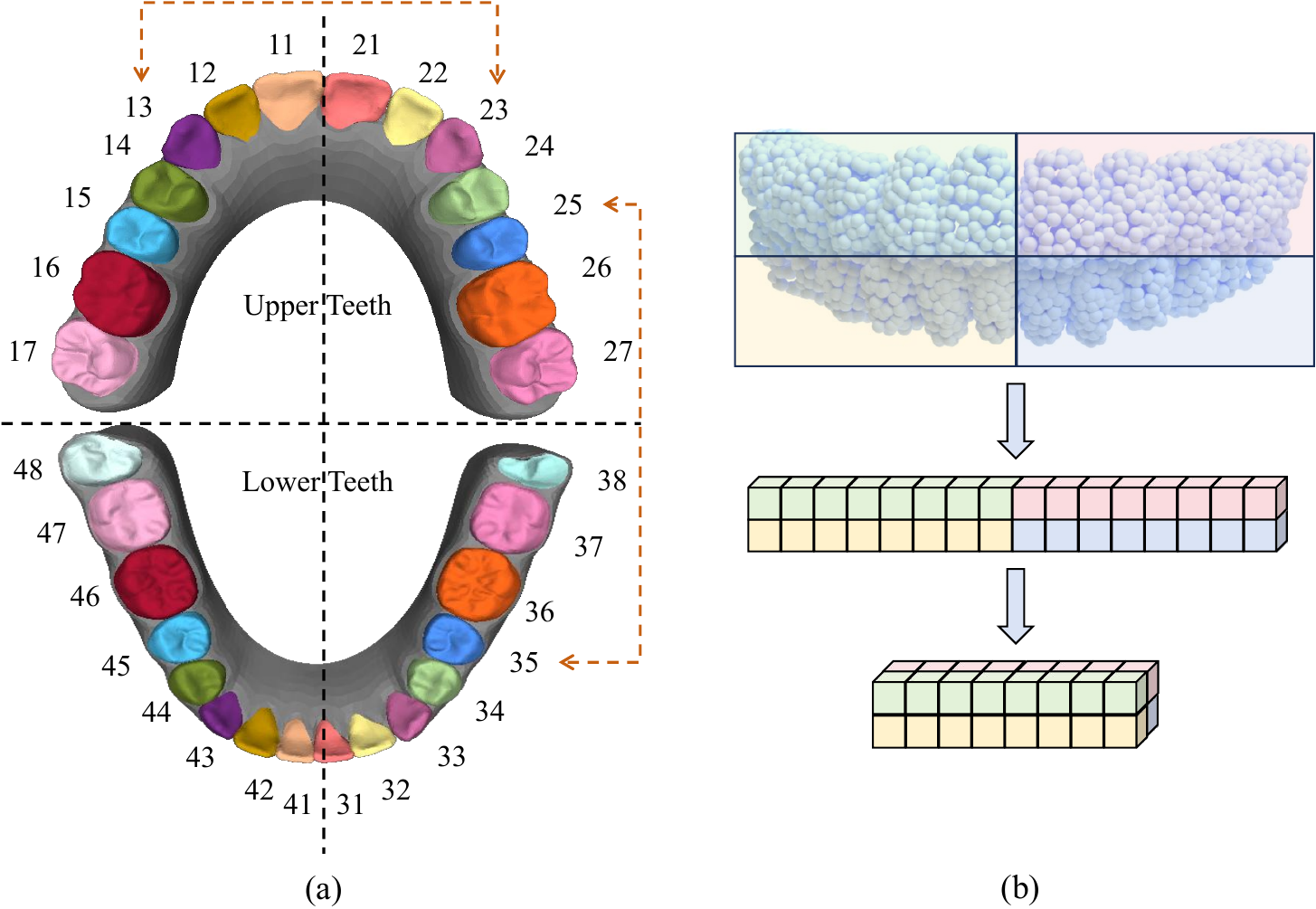}
    \caption{The spatial structure of 3D teeth models (a) and we can arrange teeth into grids of $2\times2\times8$ (b)  according to this spatial structure.}
    \label{fig:toothintroduction}
\end{figure}

\noindent
\textbf{Teeth Point Clouds Reconstruction.}
Our method aims to generate  3D teeth model $\hat{P}\in\mathbb{R}^{K\times N\times C}$, which consist of $K$ separate and closely integrated teeth point clouds, with each point cloud containing $N$ points, all in a single step. 
Considering the structural characteristics of the 3D teeth models, we propose to organize  32 teeth into a $2 \times 2 \times 8$ structured grid configuration following the Federation Dentaire Internationale (FDI) numbering system~\cite{peck1993time}, which is shown as Fig.~\ref{fig:toothintroduction}.   This spatial encoding method facilitates the computation of individual tooth features while preserving critical inter-tooth positional correlations, thereby enhancing the model's ability to learn topological relationships, 
including adjacency relationships, bilateral symmetry within the same jaw, and occlusal symmetry between opposing jaws. 
To improve the modeling of local geometric details in teeth models, we further divide the sampled 128 points of each tooth into $r^ 3$ voxels, forming a total of $[2r, 2r, 8r]$ voxels. Building upon PVCNN's paradigm~\cite{liu2019point},
we first extract internal voxel features using PointNet, followed by max-pooling operations for local feature aggregation. Subsequently, the processed voxel features are passed through cross-voxel convolution using our VQ-VAE architecture, in which both the encoder and decoder are built upon a 3D U-Net architecture. The decoder's output is fed to an MLP to get the reconstructed $\hat{P}_{post}$. 
Since the shape of $\hat{P}_{post}$ is fixed, while the number of real teeth may vary, some teeth in the output point cloud may be invalid. Consequently, the decoder's output is further passed through another MLP to generate a mask that indicates the valid teeth within $\hat{P}_{post}$.


\noindent
\textbf{Diffusion Model in Latent Space.} After training the VQ-VAE, we froze its parameters and proceeded to encode the 3D teeth point cloud into a latent space. We train the diffusion model within this latent space to learn the distribution of the latent encodings.
During sampling, noise $z$ is drawn from an isotropic Gaussian distribution that matches the dimensionality of the latent space, and then denoised by the diffusion model to obtain the latent encodings. Finally, the output of the diffusion model is passed through the VQ-VAE decoder to generate the post-orthodontic teeth model $\hat{P}_{post}$ and a mask indicating which teeth are valid in $\hat{P}_{post}$.

\subsection{Stage II: Teeth Style Generation Module}



The goal of Stage II is to generate pre-orthodontic teeth models based on the results generated by Stage I, thereby constructing a paired dataset. To enhance the capability and flexibility of our model, we aim to equip it with the ability to generate corresponding outputs conditioned on a given style model.
Specifically, we first extract individual tooth features from the given style model and feed them into the Transformer. Meanwhile, shape features extracted from the post-orthodontic model are incorporated into the attention mechanism of each Transformer layer as conditioning information. The output of the Transformer is then processed by an MLP to generate transformation parameters, which are applied to transform the post-orthodontic teeth models to exhibit the target style.
During the training stage, the post-orthodontic teeth models and the style models (pre-orthodontic) are paired samples derived from the real dataset. The style models provide the style information and serve as the ground truth simultaneously, enabling the network to effectively learn the style transformation process. During the inference stage, the post-orthodontic teeth models are generated samples from stage I, while the style model can be any arbitrary teeth model.

\begin{figure}
    \centering
\includegraphics[width=1\linewidth]{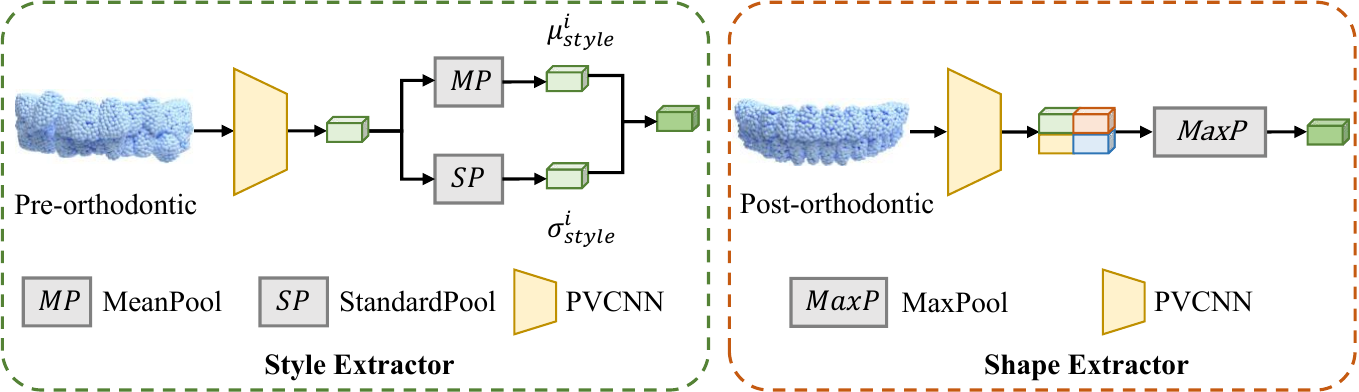}
    \caption{The framework of our proposed Style Extractor and Shape Extractor.}
    \label{fig:style}
\end{figure}

\noindent
\textbf{Style and Shape Extractor.}
As illustrated in Fig.~\ref{fig:style}, we propose a style extractor $E_{style}$ and a shape extractor $E_{shape}$ to extract style and shape information of teeth models, respectively. For $E_{style}$, we adopt the feature extraction method described in Subsection~\ref{voxel} to obtain feature maps with $r^3$ grids for each tooth. We then apply mean pooling and standard pooling along the feature dimensions to extract stylistic features. The pooled features are subsequently processed separately through two MLPs, and their outputs are added together to generate the final style encoding for each tooth. For $E_{shape}$, instead of partitioning voxels within individual teeth, we divide them across the entire teeth models, which ensures a holistic representation of the overall structure. Finally, max pooling is applied across all grids to consolidate the extracted complete shape features.

Through the aforementioned method, the extracted style information preserves fine-grained local details and distinctive stylistic features, effectively guiding each tooth toward the desired transformation. In contrast, the shape information captures the overall structure of the teeth model, serving as a conditioning input to the Transformer, ensuring sufficient context for collision avoidance.

\noindent
\textbf{Transformation Parameters Generation.}
Rather than predicting the pre-orthodontic teeth point clouds directly, we opt to predict transformation parameters $\mathbf x \in \mathbb{R}^{K\times 9}$, which could preserve tooth morphology and point correspondence between pre- and post-orthodontic teeth models.  Each $\mathbf x^i = (m^i,r^i)$ represents the transformation parameters for the $i$-th tooth of synthetic post-orthodontic point cloud $\hat{P}_{post}$, where $m_i\in{R}^3$ indicates the transition parameters for points
and  $r_i\in{R}^6$ serves as the rotation parameters for the entire tooth. Note that we compute rotation matrix $R_i\in{R}^{3\times3}$ from $r_i$ by adopting the rotation representation method described in \cite{zhou2019continuity}. Then, we apply $R_i$  to each tooth around its centroid and translate it to obtain the corresponding pre-orthodontic teeth models $\hat{P}_{pre}$.






\subsection{Loss Functions}
In \textbf{Stage I}, 
the VQ-VAE is trained to reconstruct the aggregated point clouds $\hat{P}_{post}\in\mathbb{R}^{K\times N\times C}$. 
The reconstruction loss function can be formulated as follows:
\begin{equation}
    L_{rec} = \sum_i{\mathrm{CD}(P_{i},\hat{P}_{post_{i}}}), 
    \label{CD}
\end{equation}
where  $P_{i}$ denotes the $i$-th tooth in the input point cloud $P$, $\hat{P}_{post_{i}}$ denotes the $i$-th tooth in the output, and $\mathrm{CD}$ represents the Chamfer Distance:
\begin{equation}
    \mathrm{CD}(P, Q) = \frac{1}{|P|} \sum_{p \in P} \min_{q \in Q} \| p - q \|^2 + \frac{1}{|Q|} \sum_{q \in Q} \min_{p \in P} \| q - p \|^2.
    \label{eq:chamfer}
\end{equation}

For the teeth shape diffusion model, we directly adopt Eq.~\ref{eq:diff_loss} as the loss function.


In \textbf{Stage II}, since the outputs of this stage are transformation parameters $\mathbf x \in \mathbb{R}^{K\times 9}$, we can construct the predicted pre-orthodontic teeth model $\hat{P}_{pre}$ according to $x$ and compute its distance $\mathcal{L}_{dis}$ to $P_{style}$ directly, as all points between the pre- and post-orthodontic data are corresponded.
\begin{equation}
    \mathcal{L}_{dis} = \sum_i{\Vert P_{pre} - P_{style} \Vert^2}.
\end{equation}

Following~\cite{fan2024collaborative}, we introduce a collision-avoided loss $ \mathcal{L}_{ca}$ to ensure  
that adjacent teeth remain collision-free while being as close as possible. The loss function is:
\begin{equation}
    \mathcal{L}_{ca} = \sum_{(a,b)\in\mathcal{K}}((\frac{1}{1+{d}/{s}})^{12}-2(\frac{1}{1+{d}/{s}})^{6}),
\end{equation}
where $\mathcal{K}$ represents the teeth pairs, including both adjacent teeth pairs and pairs between the upper and lower jaw. Here, $d = d_{np}+\delta$, where $d_{np}$ represents the nearest point pair distance between teeth $a$ and $b$, and $\delta$ is added to constrain $d_{np}^l$, maintaining a minimum non-overlapping distance between them.

The final loss function for stage II can be represented as:
\begin{equation}
    \mathcal{L} = \mathcal{L}_{dis} + \mathcal{L}_{ca}.
\end{equation}

\section{Experiments}
\subsection{Experimental Setups}


\noindent\textbf{Datasets.} In this paper, 
we use the 3D teeth models dataset provided in \cite{Teethdataset}, 
which is composed of 1060 pairs of pre/post-orthodontic 3D teeth models collected from real patients. After filtering out some pairs with unequal tooth counts, we obtained 720 samples for training, 80 for validation, and 120 for testing. This dataset contains samples from different categories of malocclusion, including moderate-to-severe crowding state, deep overbite, and deep overjet,  which could provide many kinds of challenging pre-orthodontic samples. 

\noindent\textbf{Training Details.}
Initially, we normalize the whole teeth model and sample 128 points from each tooth mesh using FPS~\cite{moenning2003fast} algorithm as the input. For Stage I, we configure the voxel resolution $r$ at 4  and set each voxel’s latent dimension as 64. We utilize the 3D U-Net~\cite{cciccek20163d} comprising 4 layers as the backbone for both the encoder and decoder of our VQ-VAE, as well as for the diffusion model.  We adopt Cosine schedule~\cite{nichol2021improved} as the noise schedule for our diffusion model. For Stage II, the Transformer architecture consists of 12 blocks, with 8 attention heads in each block. 
We train the proposed TeethGenerator on the platform of PyTorch in a Linux environment.
In Stage I and Stage II, we set the batch size to 32 and 64,  train the networks for 500 and 300 epochs, and use the AdamW optimizer with learning rates of 1e-3 and 1e-4, respectively.

\noindent\textbf{Evaluation Metrics for Synthetic Data.}  To evaluate the quality of our synthetic teeth models, we introduce two evaluation metrics.  
\textbf{1) Distributional Similarity:} To quantify the distributional similarity between generated teeth point clouds and the real post-orthodontic teeth point clouds,  we leverage \textit{1-NNA} (with both Chamfer distance (CD) and earth mover distance (EMD)) as the main metrics following LION~\cite{vahdat2022lion}. In Stage I, our proposed method outputs an aggregated point cloud $\hat{P}_{post}$ with a fixed shape $[32,128,3]$, along with a mask to indicate which tooth in $\hat{P}_{post}$ will be masked. For comparison, we also set the point clouds generated by baselines~\cite{yang2019pointflow,luo2021diffusion,vahdat2022lion,mo2023dit,zhou20213d} to contain 4096 points.
   \textbf{2) Uniqueness:} Moreover, we aim to quantify the uniqueness of the synthetic data. Similar to ~\cite{kim2023dcface}, we define the  metric $U_{CD} =\frac{\left| S \right|}{N}$  to reflect the ratio of unique samples, where  $\left| S \right|$ represents the number of unique samples in the dataset, which is defined as:
\begin{equation}
    S = \{f_i:\mathrm{CD}(f_i,f_j)>r,j<i,i,j\in\{1\dots N\}\}
    \label{S}
\end{equation}
where $\mathrm{CD}$ denotes Chamfer Distance as defined in Eq.~\ref{eq:chamfer}, $f_i$ denotes the $i$-th sample in the dataset and N denotes the size of the dataset. In this paper, we set the threshold $r$ as 1 cm.




\noindent\textbf{Mesh Reconstruction for Visualization.} To provide a clearer visualization of teeth arrangement, we convert the point clouds into meshes based on our real dataset. Specifically, for each tooth in the synthetic point cloud $\hat{P}$, we randomly select an initial mesh with the corresponding FDI notation label. We then perform non-rigid registration to align the selected mesh with the source point cloud.

\subsection{3D Teeth Generation Performance}

\begin{figure*}[t]
    \centering
    \includegraphics[width=0.92\linewidth]{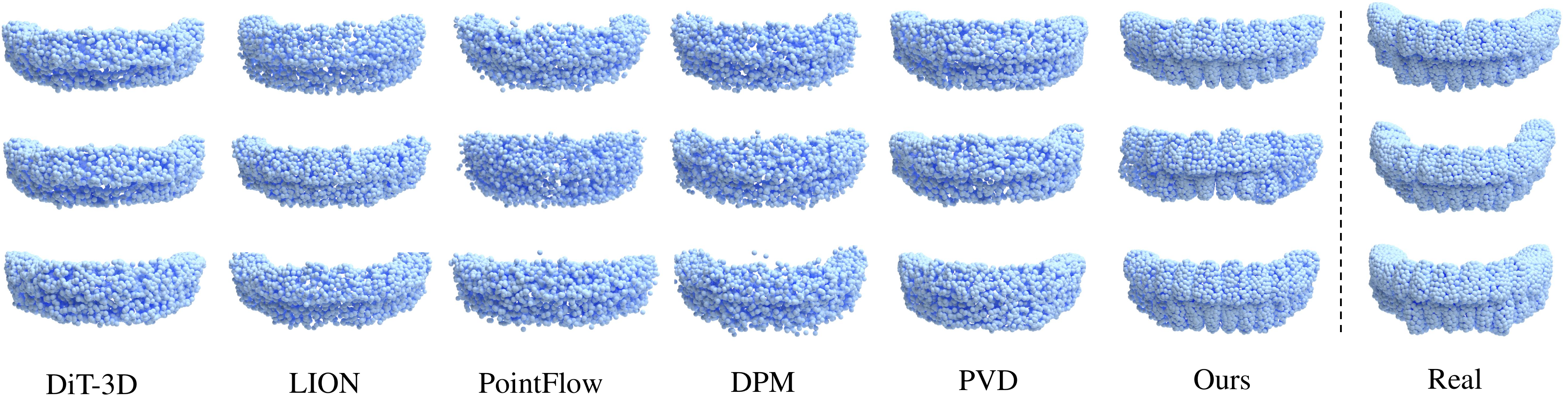}
    \caption{Visualization of the post-orthodontic teeth models generated by state-of-the-art general 3D shape generation methods and our methods. To provide clear and reasonable tooth morphology, we present three randomly selected real teeth models in the last column.  }
    \label{fig:compare}
\end{figure*}

\begin{figure*}[t]
  \centering
  \begin{minipage}[b]{0.4\textwidth} 
    \centering
    \includegraphics[width=0.9\linewidth]{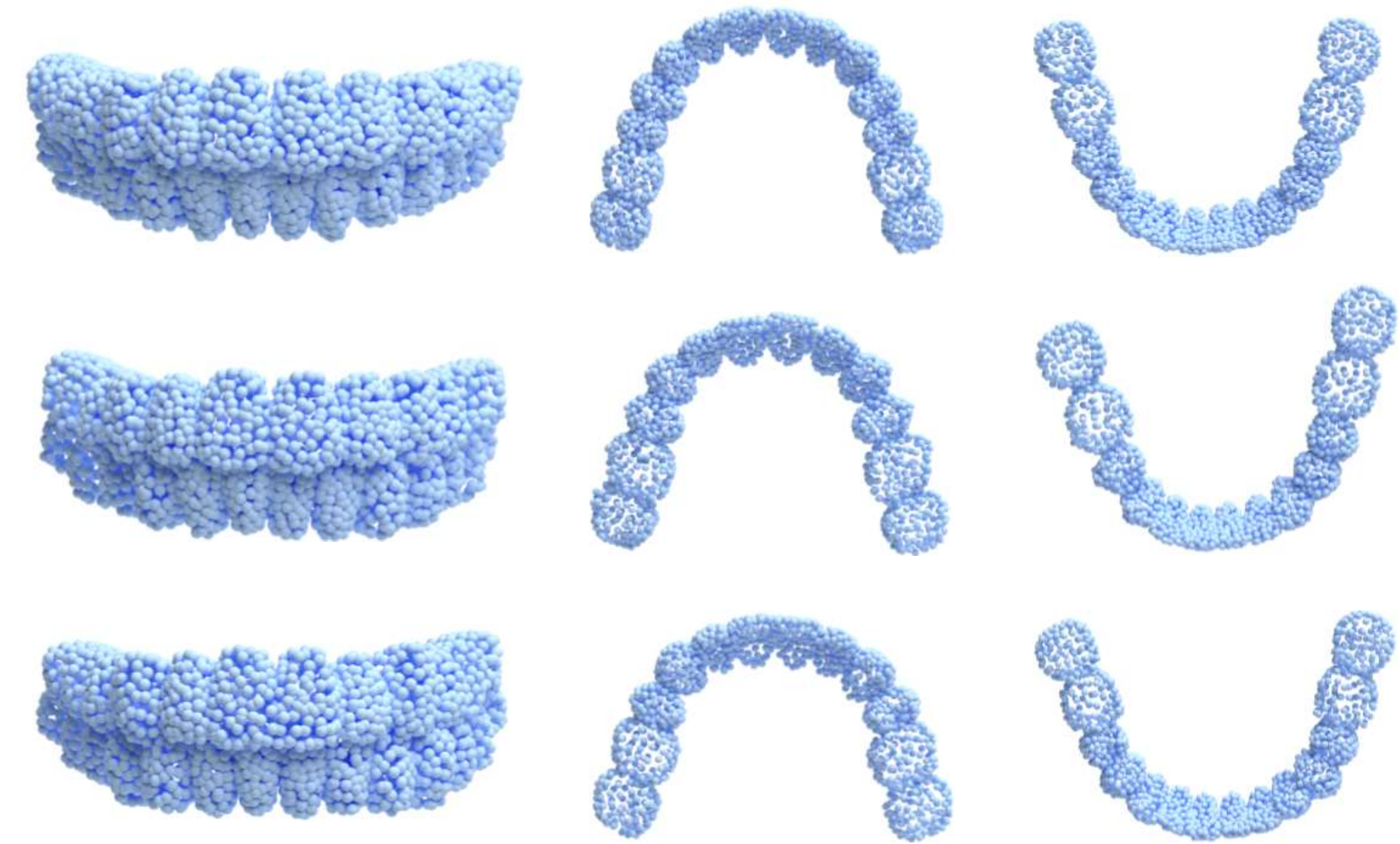} 
    \caption{Generated samples from Stage I are presented. Each row displays an aggregated teeth model, with columns representing the full model, upper arch, and lower arch, respectively.}
    \label{fig:samples}
  \end{minipage} 
\hspace{1mm}
  \begin{minipage}[b]{0.55\textwidth} 
    \centering
    \includegraphics[width=1\linewidth]{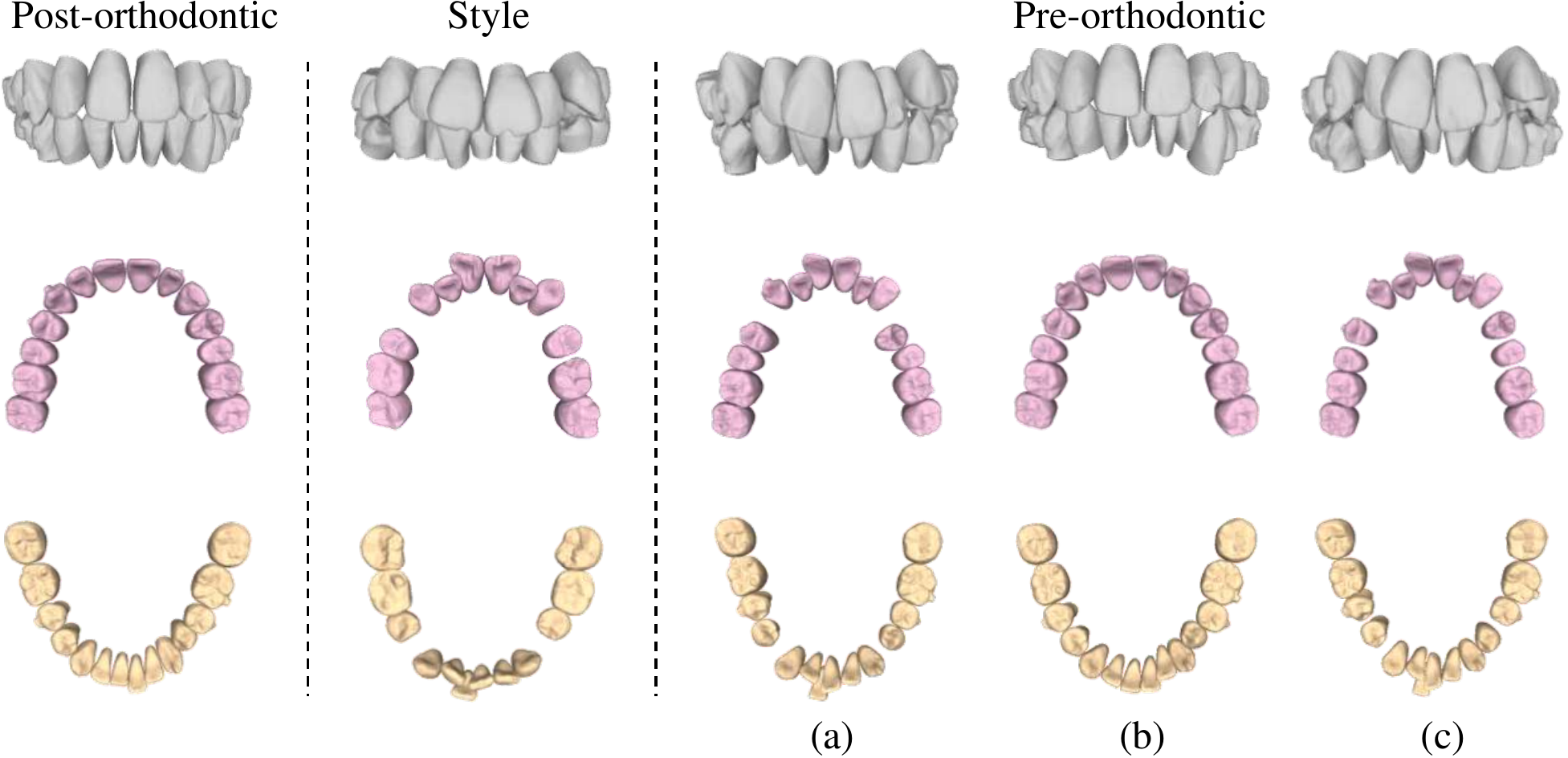} 
    \caption{For the same post-orthodontic teeth model and style model, (a) presents the result of our proposed method, (b) shows the outcome when the voxelization strategy is swapped, and (c) illustrates the result of the straightforward approach.}
    \label{fig:ablation2}
  \end{minipage}
\end{figure*}

\noindent
\textbf{Post-orthodontic 3D Teeth Generation.}
To validate the effectiveness of TeethGenerator, we compare it with several state-of-the-art \textbf{general} point cloud generation methods, i.e. PointFlow~\cite{yang2019pointflow}, DPM~\cite{luo2021diffusion}, LION~\cite{vahdat2022lion}, 
PVD~\cite{zhou20213d} and DiT-3D~\cite{mo2023dit}. The quantitative results are summarized in Table~\ref{tab:comparison}, where our method achieves the best performance. In Fig.~\ref{fig:compare}, we illustrate the visualization results of the teeth point clouds generated by comparison methods and our method (Stage I). As can be seen, existing methods fail to construct precise teeth shapes and are unable to segment teeth models into independent teeth. Furthermore, we demonstrate the upper and lower arches of our generated teeth point clouds in Fig.~\ref{fig:samples}. Our results feature diverse teeth morphologies, clear boundaries between individual teeth, and varying numbers of teeth, which closely resemble the characteristics of real teeth datasets.

\begin{figure*}[t]
    \centering
    \includegraphics[width=0.92\linewidth]{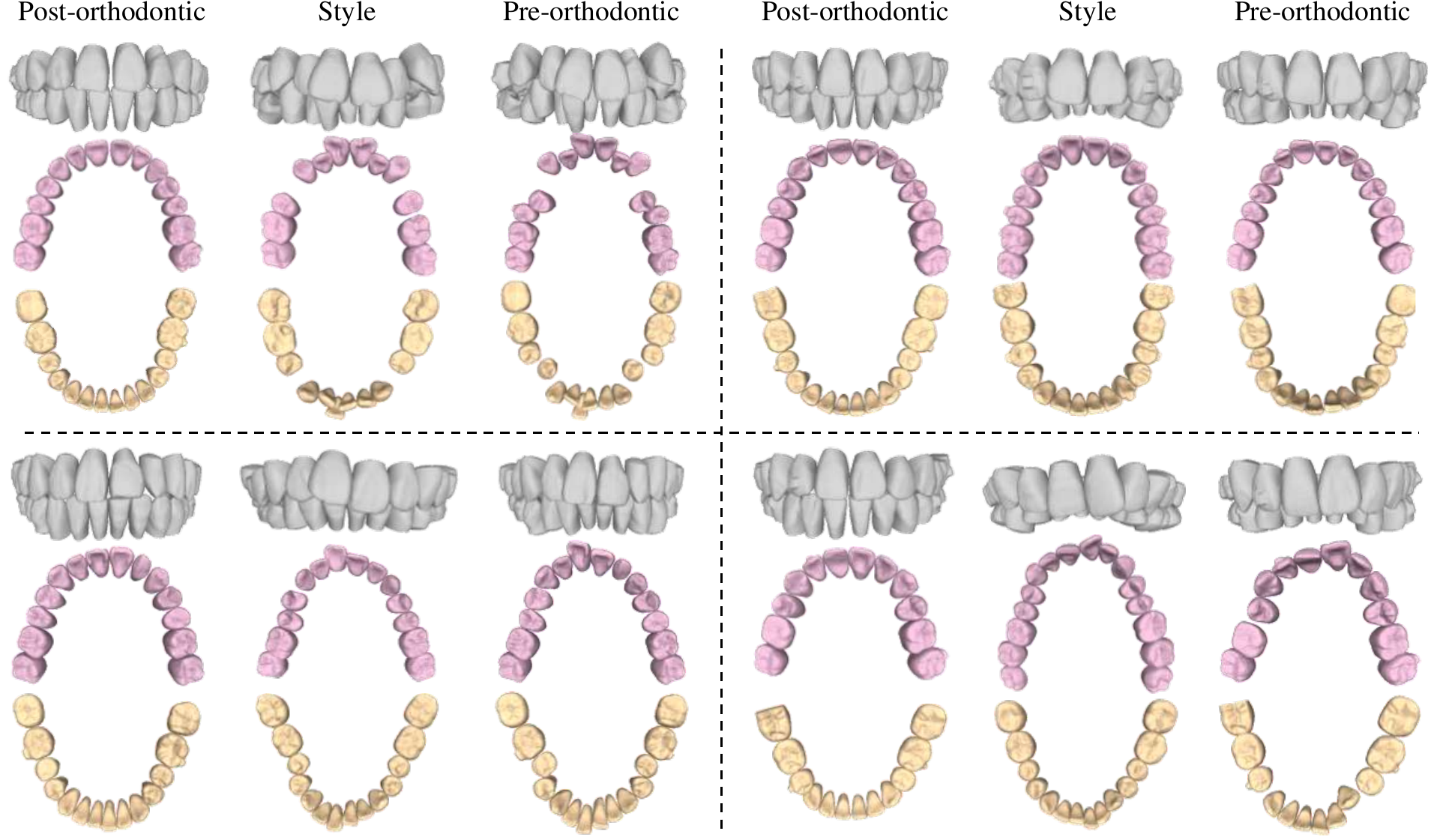}
    \vspace{-3pt}
    \caption{Visualization results of generated pre-orthodontic teeth models conditioned on diverse style models.  }
    \label{fig:transfer}
    \vspace{-6pt}
\end{figure*}

\noindent
\textbf{Pre-orthodontic 3D Teeth Generation.}
Here, we demonstrate the generation performance of pre-orthodontic 3D teeth models in Fig.~\ref{fig:transfer}. It can be observed that the proposed Teeth Style Generation Module effectively captures the characteristics in the style models and successfully applies them to alter the arrangements of the post-orthodontic models, achieving the realistic synthesis of pre-orthodontic teeth models.  The consistent performance across diverse style variations demonstrates the flexibility and generalization capability of our framework.


\begin{table}[t]
\resizebox{\linewidth}{!}{
\centering
\begin{tabular}{lccc}
\toprule
\textbf{Model} & \textbf{CD} (\%, $\downarrow$) & \textbf{EMD} (\%, $\downarrow$) & $\mathbf{U_{CD}}$ (\%, $\uparrow$) \\ \midrule

PointFlow\cite{yang2019pointflow} & 97.62 & 83.88 & 62.22 \\
DPM\cite{luo2021diffusion} & 89.25 & 74.50 & 75.69 \\
PVD\cite{zhou20213d} & 84.87 & 78.12 & 49.58\\
\midrule
LION\cite{vahdat2022lion} & 90.41 & 77.93 & 52.78 \\
DiT-3D\cite{mo2023dit} & 95.75 & 82.01 & 34.03 \\
TeethGenerator & \textbf{69.50} & \textbf{71.88} & \textbf{96.25} \\

\bottomrule
\end{tabular}}
\caption{Results (1-NNA and uniqueness) of generated teeth models, where the metrics are  calculated on 720 generated samples.}
\label{tab:comparison}
\end{table}




\subsection{Effectiveness for Downstream tasks}
\label{sec:downstream}

Our goal in generating synthetic 3D teeth models is to create more complex samples that can benefit the training of the tooth alignment networks.
To further evaluate the effectiveness of our synthetic data on downstream tasks, we conduct teeth arrangement experiments, comparing results using real data alone as well as a combination of real and synthetic data.  In this subsection, we employ TANet~\cite{wei2020tanet} as the backbone, an automatic teeth alignment model that relies exclusively on point cloud data.

\begin{figure}
    \centering
\includegraphics[width=1.0\linewidth]{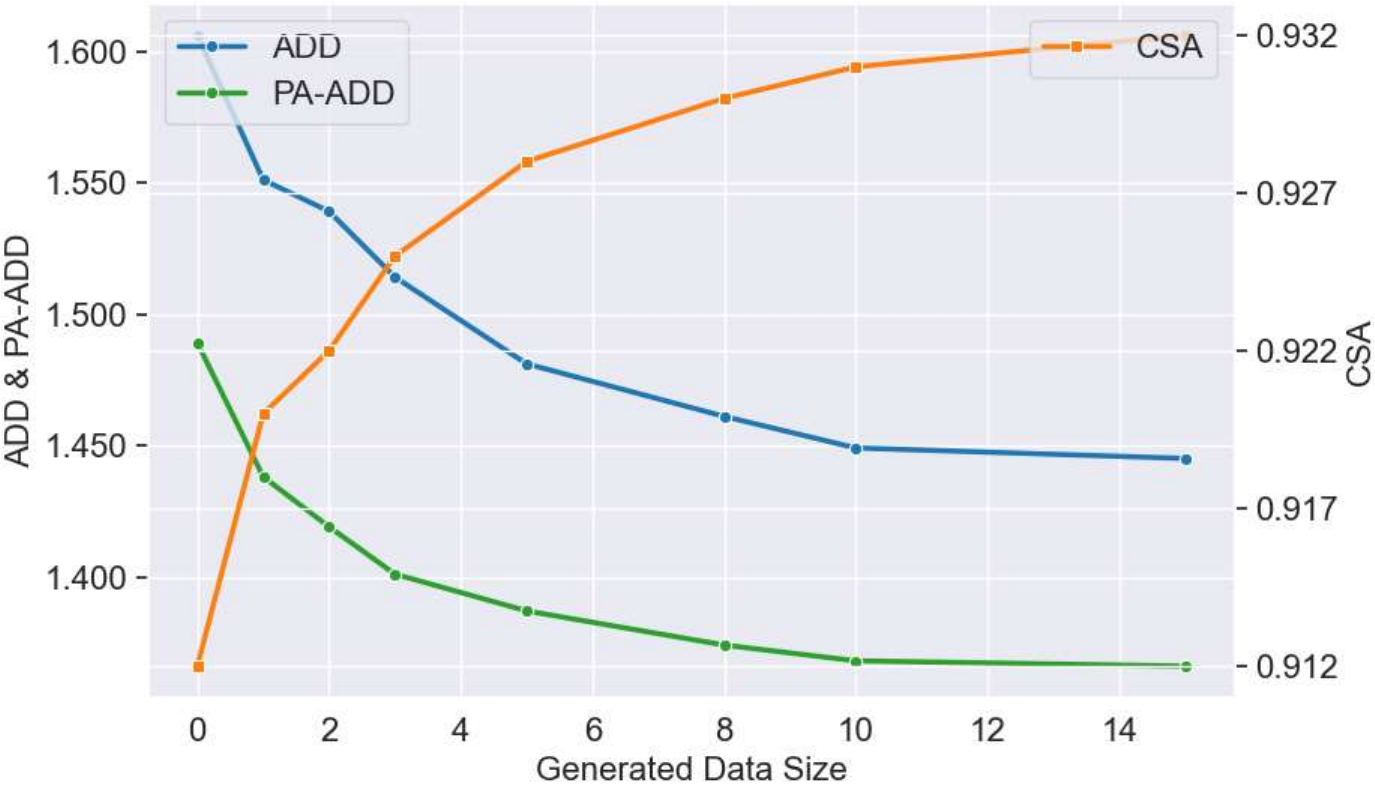}
    \caption{Improved tooth alignment performance of TANet as the number of synthetic teeth models increases. }
    \vspace{-2pt}
    \label{fig:datasize}
\end{figure}

In the real dataset~\cite{Teethdataset}, the training set consists of 720 teeth models. We first randomly select the style models from the real pre-orthodontic data to generate 720 synthetic teeth models, matching the size of the real training set. Then, we combine the real and synthetic data to train the TANet.  The experiment results are shown in Fig.~\ref{fig:datasize}, where ADD, PA-ADD, and CSA  are three common tooth alignment evaluation metrics following TANet. It is noted that the generated data size, $n$, represents how many times the training set is synthesized. Specifically, $n=0$  refers to training the network with the real dataset only, while the other results involve training the network by merging the real dataset and synthetic dataset with the quantity of '$n\times720$'.
Fig.~\ref{fig:datasize} shows that performance continues to improve as the amount of generated data increases up to 10 times the amount of real data, and then gradually converges.
This experiment fully demonstrates the significance of our synthesized data, which can promote the performance of downstream tooth arrangement tasks.






\subsection{Ablation studies}



\noindent
\textbf{Voxelization Strategies in Stage I.}
In Stage I, we performed an ablation analysis to evaluate different voxelization strategies and the organization of tooth structure. The quantitative comparisons are listed in Table~\ref{tab:ablation}.  According to the results, we draw the following key conclusions: \textbf{(i)} Larger voxel sizes enhance geometric detail capture at the cost of cubic growth in memory consumption. When  $r=5$, TeethGenerator may struggle to handle the increased volume of information. \textbf{(ii)}  The significant performance degradation in No. 4 indicates that voxel-based feature extraction is crucial in our framework. \textbf{(iii)} Global voxel partitioning (No. 5) underperforms localized tooth-wise partitioning, as some voxels contain points from two different teeth or gaps between the teeth,  leading to inefficient feature extraction.
\textbf{(iv)} Compared with simply concatenating the tooth point cloud into a $1\times1\times32$ matrix (No. 6), the proposed space structure shown in Fig.~\ref{fig:toothintroduction}(b), could explicitly model bilateral symmetry within jaws and occlusal relationships between opposing jaws, which enable a better interaction learning.

\begin{table}[h]
\resizebox{\linewidth}{!}{
\centering
\begin{tabular}{cccccc}
\toprule
No.&Space Structure&Voxelize & \textbf{CD}$\downarrow$  & \textbf{EMD} $\downarrow$ & $\mathbf{U_{CD}} \uparrow$ \\
\midrule

1&$2\times2\times8$&$3\times3\times3$   & 82.63 & 81.50 & 92.64 \\
2&$2\times2\times8$&$4\times4\times4$  & \textbf{69.50} & \textbf{71.88} & \textbf{96.25}  \\
3&$2\times2\times8$&$5\times5\times5$   & 77.75 & 83.23 & 95.56\\
4&-&no  & 89.59 & 84.13 & 21.11 \\
5&-  &$8\times8\times32$ & 78.16 & 76.91 & 89.03 \\
6&$1\times1\times32$ &$4\times4\times4$ & 74.48 & 72.82 & 94.17 \\
\bottomrule
\end{tabular}}
\caption{Results (1-NNA and uniqueness) of generated teeth models with different voxel sizes and network structure. 
In No. 4, teeth-wise features are encoded by PointNet and then fed into a Transformer to capture the interaction information between teeth.
}
\label{tab:ablation}
\end{table}

%
\noindent
\textbf{Style Transfer Strategies in Stage II.}  
We compare our method with another two style transfer strategies. 
\textbf{(i)} Since the ground truth transformation parameters are known,
a straightforward approach is to apply these parameters to the change the pose of post-orthodontic teeth models.
However, due to teeth amounts and sizes between the post-orthodontic teeth  model and the style  model may differ,  directly applying the ground truth parameters becomes unrealizable. 
%
Moreover, the straightforward method requires paired data to compute ground truth parameters. In contrast, our proposed method only takes the style model as the condition, eliminating the need for paired data. 
\textbf{(ii)} Besides,  we also conducted experiments by swapping the voxelization strategies of $E_{style}$ and $E_{shape}$ to explore the dependency between style information and local details. Specifically, we partition voxels across the entire teeth model for $E_{style}$ and within individual teeth for $E_{shape}$. 
Visualization results of the generated pre-orthodontic teeth model are shown in Fig.~\ref{fig:ablation2}, where (a) presents the results of our method. (b) shows the outcome when the voxelization strategy is swapped, where the final output fails to replicate the desired style due to $E_{style}$ cannot extract  sufficient local details. (c) illustrates the result of the straightforward approach, which can be observed obvious gaps between teeth.
\section{Conclusion}


This paper presents a two-stage teeth data synthetic framework to generate
the pre- and post-orthodontic teeth models. By learning the distribution of tooth morphology and integrating the desired teeth style and shape information, our method could synthesize paired teeth models  consistent with the true distribution.
Extensive experimental results demonstrate the proposed method achieves state-of-the-art data generation performance. Moreover, we explore the benefits of using synthetic data for training in downstream tasks, further verifying the effectiveness of the proposed data synthesis method.

{
    \small
    \bibliographystyle{ieeenat_fullname}
    \bibliography{main}
}
\end{document}